%% file: sn-article.tex
\documentclass[pdflatex,sn-mathphys-num]{sn-jnl}

\usepackage{graphicx}%
\usepackage{multirow}%
\usepackage{amsmath,amssymb,amsfonts}%
\usepackage{mathrsfs}%
\usepackage[title]{appendix}%
\usepackage{xcolor}%
\usepackage{textcomp}%
\usepackage{manyfoot}%
\usepackage{booktabs}%
\usepackage{algorithm}%
\usepackage{algorithmicx}%
\usepackage{algpseudocode}%
\usepackage{listings}%
\usepackage{listings}
\usepackage{color}
\usepackage{xcolor}
\usepackage{graphicx}
\usepackage{float}
\usepackage{physics}
\usepackage{comment}
\usepackage{tcolorbox}
\usepackage{rotating} 
\usepackage{bbm} 
\usepackage{upgreek} 

\definecolor{dkblue}{rgb}{0,0.39,0}
\definecolor{gray}{rgb}{0.66,0.66,0.66}
\definecolor{mauve}{rgb}{0.91,0.33,0.50}
\definecolor{gold}{rgb}{1,0.84,0}

\usepackage{amsthm}%
\begin{document}

\title{Achieving Optimal Tissue Repair Through MARL with Reward Shaping and Curriculum Learning}

\author*[1]{ \fnm{Muhammad Al-Zafar} \sur{Khan}}\email{Muhammad.Al-ZafarKhan@zu.ac.ae}

\author[1,2]{\fnm{Jamal} \sur{Al-Karaki}}\email{Jamal.Al-Karaki@zu.ac.ae}

\affil*[1]{\orgname{College of Interdisciplinary Studies}, Zayed University, \state{Abu Dhabi}, \country{UAE}}

\affil[2]{\orgname{College of Engineering}, The Hashemite University, \state{Zarqa}, \country{Jordan}}

\abstract{In this paper, we present a multi-agent reinforcement learning (MARL) framework for optimizing tissue repair processes using engineered biological agents. Our approach integrates: (1) stochastic reaction-diffusion systems modeling molecular signaling, (2) neural-like electrochemical communication with Hebbian plasticity, and (3) a biologically informed reward function combining chemical gradient tracking, neural synchronization, and robust penalties. A curriculum learning scheme guides the agent through progressively complex repair scenarios. In silico experiments demonstrate emergent repair strategies, including dynamic secretion control and spatial coordination.}
\keywords{Intelligent Tissue Repair, Multi-Agent Reinforcement Learning (MARL), Computational Biology}

\maketitle

\section{Introduction}\label{introduction}
Tissue repair and regeneration represent one of the most fundamental and miraculous biological processes critical to organism survival and longevity. The complex, self-organizing mechanisms through which damaged tissues restore functionality involve intricate coordination among diverse cell populations, extracellular matrix components, signaling molecules, and mechanical forces. The work of Ross Granville Harrison (1870--1959), where he grew frog embryonic tissue outside the body, laid the foundation for tissue engineering as a formalized discipline. Later work by Robert Langer and Joseph Vacanti in the later 1980s and early 1990s propelled the field into its modern form by creating artificial scaffolds made of biocompatible polymers and seeding them with living cells culminating in the famous \textit{Vacanti Mouse} experiment where cartilage tissue in the shape of a human ear was grown on a mouse's back \cite{cao1997transplantation}.

Despite significant advances in regenerative medicine and tissue engineering, our ability to precisely control and accelerate tissue repair processes remains limited, particularly in severe trauma, chronic wounds, degenerative conditions, and age-related tissue decline. This limitation stems largely from the staggering complexity of biological repair systems that operate across multiple spatial and temporal scales, involving countless parallel and sequential cellular decisions that collectively drive tissue restoration.

Recent breakthroughs in AI, particularly in Reinforcement Learning (RL) -- where we have observed its effectiveness in large search space games like Go \cite{silver2016mastering,silver2017mastering}, offer a potential opportunity to address this complexity. Multi-Agent RL (MARL) involves several autonomous agents that work together cooperatively, competitively, or in a mixed-motive setting, interact with the environment, pursue individual objectives, and contribute towards maximizing the reward of the entire system. The MARL paradigm naturally maps onto biological tissue repair processes, where different cell types -- including fibroblasts, macrophages, endothelial cells, and stem cells -- act as independent yet cooperative agents within the wound microenvironment. Each cell type senses local conditions, executes specialized functions, adapts to changing circumstances, and communicates with neighboring cells to orchestrate tissue regeneration. The emergent collective behavior of these cellular agents ultimately determines healing outcomes, from successful restoration of tissue architecture and function to pathological states such as fibrosis or chronic inflammation.

While traditional MARL approaches have shown considerable promise in various domains, including robotics \cite{busoniu2008comprehensive,singh2022reinforcement,orr2023multi}, transportation networks and autonomous vehicles \cite{haydari2020deep}, and resource management \cite{cui2019multi}, their direct application to biological systems faces significant challenges. Biological environments exhibit:
\begin{enumerate}
\item \textbf{Extreme Heterogeneity:} Biological environments exhibit a vast degree of diversity and variability on multiple scales that arise on the cellular, molecular, structural, and temporal levels. 
\item \textbf{Partial Observability:} Localized cell groups within the microstructure have a limited view of the overall system and are confined to paracrine and juxtacrine signaling, confined mechanical cues, and limited diffusion. 
\item \textbf{Asynchronous Action Execution:} Different biological components -- such as cellular processes, molecular interactions, and external stimuli -- within a tissue repair environment operate on different time scales and execute actions nonparallel. 
\item \textbf{Delayed Reward Features:} The outcomes of actions taken by biological components during tissue repair often manifest with significant delays. For example, the consequences of one molecular interaction can trigger a cascade of events that unfold over time. 
\end{enumerate}
These inherent factors can destabilize the direct application of conventional RL algorithms. Furthermore, the sheer number of interacting agents in tissue repair -- which constitute millions of cells -- far exceeds the scale of most current MARL implementations. Perhaps most critically, the reward signals that guide learning in biological systems remain poorly understood, as cells must balance immediate local objectives against long-term tissue-level goals without centralized coordination.

Reward shaping, which is the strategic modification of reward functions to facilitate learning, emerges as one powerful yet simple technique to address these biological complexity challenges. Designing biologically inspired reward structures that incorporate domain knowledge about tissue homeostasis, developmental principles, and healing trajectories makes it possible to design multi-agent tissue repair systems that can potentially guide optimal repair strategies. Properly shaped rewards can encode both local cellular imperatives -- such as survival, differentiation, and migration -- and global tissue objectives --  such as restoration of mechanical integrity, vascularization (the formation of blood vessels), and functional recovery -- enabling agents to learn effective policies across spatial and temporal scales. Moreover, reward shaping can accelerate learning in sparse-reward environments characteristic of biological systems, where meaningful feedback may only become available after extended sequences of cellular actions.

This paper presents a framework integrating MARL architectures with biologically informed reward shaping and curriculum learning to model and optimize tissue repair processes. The objective of this research is to present an intelligent process of tissue repair in regenerative medicine. Through this approach, we aim to demonstrate that appropriately designed MARL systems with biologically informed reward shaping can capture the emergent intelligence of tissue repair processes and can reveal an optimized healing strategy beyond the current biological understanding. We believe that our original contribution with this work presents a step towards developing truly intelligent therapeutic interventions capable of guiding and accelerating tissue repair in situations ranging from chronic wounds to degenerative conditions, with the ultimate goal being to improve human health and longevity. 

Our approach is as follows: The system consists of engineered biological agents that learn cooperative strategies using MARL to achieve the collective goal of repairing a tissue site. In order to guide the learning process, we incorporate reward shaping that recompenses long-term success by:
\begin{enumerate}
\item Rewarding cells for migrating towards the wound site/s.
\item Rewarding the production of specific growth factors known to promote healing.
\item Penalizing actions that lead to inflammation or tissue damage.
\end{enumerate}
Further, we incorporate curriculum learning in order for the learning process to go step-by-step and increase in complexity gradually from elementary to more sophisticated, as opposed to the agent being immediately exposed to the full complexity of the problem from an early stage.

We believe this approach is adapted specifically for the unique challenges inherent in biological systems. By posing the task of tissue repair as a MARL problem with carefully designed reward structures, we aim to uncover emergent repair strategies that may not be immediately evident through traditional biological investigation alone.

This paper is divided as follows:

In Sec. \ref{related work}, we discuss some parallel work to our research.

In Sec. \ref{the approach}, we outline the methodology we adopted in training the model.

In Sec. \ref{experiment}, we present the results of our experiments and provide a discussion on their relevance.

In Sec. \ref{conclusion}, we present closing remarks reflecting on what was achieved and how the work in this paper can be extended. 

\section{Related Work}\label{related work}
The work in \cite{gharibshahian2024recent} is a rigorous review of the various AI methods applied for intelligent tissue repair. Within the context of robotic platforms and automated systems, which our work falls under, the authors highlight several important papers where scaffolds were used as robots and computational optimization was applied to achieve optimal tissue repair. Other works, such as in \cite{bagherpour2025application,liu2025current}, emphasize supervised and unsupervised learning applications in diagnoses and classifications, with little to discussion points on the direct application of RL agents performing repairs. An alternative to these is the work in \cite{datta2021reinforcement} which discusses the applications of how RL is used in surgical techniques. A brief discussion of some recent work on the applications of RL in surgery is discussed below.

\cite{lu2024accelerating}, the authors present a model-free approach that combines deep RL (DRL) and model control using bioelectronic sensors and actuators to adaptively control wound healing in real-time. Both in simulation and real-world applications, the method demonstrated promising results with the potential for clinical use cases. 

In \cite{shahkoo2023autonomous}, the authors use DRL with an evolutionary algorithm to learn the optimal tensioning policy in order to accurately cut deformable soft tissue using a robot equipped with surgical scissors and a gripper. The results showed that the method achieved less damage, more accurate cuts, and higher performance scores than previous methods.

Many other studies discuss surgical techniques, the healing process, and learning optimal actions for operative procedures. However, the gap in the literature is that there are virtually no investigations of using bioengineered intelligent agents to perform surgical procedures at repair sites in the body. Our research aims to fill this gap by presenting a possible approach that can be built upon.  

\section{The Multi-Agentic Approach}\label{the approach}
We model the system as follows:
\begin{itemize}
\item The \textbf{agents} consist of engineered cells or bacteria that are injected or orally ingested into the patient's body.
\item The \textbf{states} consists of the localized molecular concentrations and the internal health metrics at the repair sites of the patient.
\item The \textbf{actions} consist of the agents' secreting repair factors, moving towards the repair sites, and amplifying signals.
\item The \textbf{rewards} are the biochemical signals such as adenosine triphosphate (ATP) that are used and stored at the cellular level, and isopropyl $\beta$-D-1 thiogalactopyranoside (IPTG) used to cause conformational changes by binding to the bind lac repressor proteins. 
\end{itemize}
The communication between agents occurs via three parallel channels:
\begin{itemize}
\item \textbf{Diffusion-Based Chemical Signaling:} The concentration of molecule $i$ follows stochastic reaction-diffusion equations
\begin{equation}\label{chemical signaling eqn}
\frac{\partial C_{i}}{\partial t}=D_{i}\nabla^{2}C_{i}-\lambda_{i}C_{i}+\sum_{j}S_{j}(\mathbf{x},t)+\eta_{i}(\mathbf{x},t),
\end{equation}
where for molecule $i$, $C_{i}(\mathbf{x},t)$ is the concentration at position $\mathbf{x}$ and time $t$, $D_{i}$ is the diffusion coefficient, $\lambda_{i}$ is the concentration decay constant, $S_{j}$ is the source terms (including sinks), and $\eta$ is the spatiotemporal Gaussian (white) noise term with the property that its average $\langle\eta_{i}(\mathbf{x},t)\eta_{j}(\mathbf{x}',t')\rangle=\sigma^{2}\delta_{ij}\delta(\mathbf{x}-\mathbf{x}')\delta(t-t')$, with $\sigma$ accounting for thermal and biological variability, $\delta_{ij}$ being the Kronecker delta function, and $\delta(\mathbf{x}-\mathbf{x}')$ and $\delta(t-t')$ being the spatial and temporal Dirac delta functions, respectively. The model in \eqref{chemical signaling eqn} is substantiated in the works of \cite{turing1990chemical} with explicit models found in, for example, \cite{elf2003mesoscopic}.
\item \textbf{Synaptic/Electrochemical Neural-Like Signaling:} Short-range fast communication occurs via gap junctions and nanowires in bacterial nanotubes. From agent $p$ to agent $q$, we can express the time-dependent signal as
\begin{equation}\label{synaptic signaling eqn}
I_{pq}(t)=w_{pq}(t)\mathcal{A}[a_{p}(t)],
\end{equation}
where $I_{pq}$ is the transmitted signal between agents $p$ and $q$, $w_{pq}$ is the connection strength learned through Hebbian learning and $\mathcal{A}$ is the neural activation function with membrane potential $a_{p}$ that acts as the actions taken by the agent. The equation in \eqref{synaptic signaling eqn} is substantiated by standard models in Computational Neuroscience like the Hodgkin-Huxley model, the Wilson-Cowan model for rate-based neurons, the various connectionist models (artificial neural networks), and Hebbian learning theory.  
\item \textbf{Stochastic Noise:} This term accounts for environmental variability that occurs through random (Brownian) motion and molecular degradation, and makes the model more realistic. Thus, the total input to agent $q$ is simply the sum total of all signals
\begin{equation}
h_{q}=\sum_{p}I_{pq}(t)+\mathcal{C}[C_{i}(\mathbf{x}_{q},t)],    
\end{equation}
where $\mathcal{C}$ is the chemical sensing term that arises from probing the concentration of molecule $i$. 
\end{itemize}
We assume that the dynamics of each state follows
\begin{equation}
\mathbf{s}_{k}(t+1)=f(\mathbf{s}_{k},a_{k},\mathbf{a}_{-k})+\xi_{k}(t),
\end{equation}
where $\mathbf{s}_{k}(t)$ is the joint states at time $t$, $\mathbf{a}_{k}$ are the joint actions in the previous state, $f$ is the model of the state dynamics, and $\xi_{k}$ are the noise associated with metabolic fluctuations. Agents sample actions from a noisy policy distribution
\begin{equation}\label{action sampling eqn}
a_{k}(t)\sim\pi_{k}(\mathbf{s}_{k})+\zeta_{k}(t),
\end{equation}
where $\zeta_{k}$ is the action noise associated with the Brownian motion in chemotaxis. Unlike traditional RL that only accounts for scalar signals, we extend our reward function to incorporate chemical cooperation, neural synchronization, and a robustness penalty according to
\begin{equation}\label{reward function}
R_{k}(t)=R_{\text{ext}}(t)++\beta_{1}r_{\text{chem}}+\beta_{2}r_{\text{neu sync}}(t)+\beta_{3}r_{\text{robust}}(t),
\end{equation}
where 
\begin{equation}\label{reward shaping}
\begin{aligned}
r_{\text{chem}}(t)=&\;\sum_{i}\left[C_{i}(\mathbf{x}_{k},t)-C_{i}(\mathbf{x}_{\text{inj}},t)\right]^{2}, \\ 
r_{\text{neu sync}}(t)=&\;\sum_{q}\left[a_{k}(t)-a_{q}(t)\right]^{2}, \\
r_{\text{robust}}(t)=&\;-\text{Var}(a_{k}), 
\end{aligned}
\end{equation}
where $\mathbf{x}_{\text{inj}}$ denotes the position of the injury site, and this mean-squared error (MSE) form of the chemical cooperation encourages gradient following, the neural synchronization reward promotes coordinated firing, and the robustness penalty discovers noise-tolerant policies, with $0\leq\beta_{i}\leq 1$ being the neural coefficients. The objective function is optimized by applying the policy gradient method that optimizes chemical and neural policies via
\begin{equation}
\nabla_{\theta_{k}}J(\theta_{k})=\mathbb{E}\left[\nabla\pi_{k}(a_{k}|\mathbf{s}_{k})\cdot A_{k}\right]+\mu\nabla H(\pi_{k}),    
\end{equation}
where $H$ is the regularized entropy that encourages the agents to explore in the biological environment, which is noisy, $A_{k}$ is the advantage computed via a centralized critic which accounts for neural coupling, and $\mu$ is a weighting parameter.   

The curriculum aspect is integrated into the learning process in a linear manner in which the agents initially start off with elementary tasks that steadily increase in difficulty according to
\begin{equation}\label{curriculum eqn}
\mathcal{T}(t)=\mathcal{T}_{0}+\left(\mathcal{T}_{f}-\mathcal{T}_{0}\right)\min\left(\frac{t}{n},1\right),
\end{equation}
where $\mathcal{T}$ is the target value being gradually adjusted, $\mathcal{T}_{0}$ is the value at the beginning of the curriculum, $\mathcal{T}_{f}$ is the learning value that the agent wants to reach, and $n$ is the total number of iterations. 

We summarize the process of training agents to perform intelligent tissue repair in Algorithm \ref{algo1} below.

\begin{algorithm}[H]
\caption{Smart Tissue Repair with MARL}
\label{algo1}
\begin{algorithmic}[1]
\State \textbf{input} $N$ agents with policies $\pi_{k}$, chemical parameters ($D_{i},\lambda_{i},\sigma$), neural parameters ($0\leq\beta_{1:3},\mu\leq1$), curriculum targets ($\mathcal{T}_{0},\mathcal{T}_{f},n$), activation $\mathcal{A}$, learning rate $0\leq\alpha\leq 1$, decay rate $0\leq\gamma\leq 1$, total time steps $T\in\mathbb{Z}^{+}$
\State \textbf{initialize} concentration field $C_{i}(\mathbf{x},0)=0$, neural weights $w_{pq}\sim\mathcal{N}(0,1)$, target $\mathcal{T}(0)=\mathcal{T}_{0}$, target repair site $\mathbf{x}_{\text{inj}}$
\For {$t$ from $0$ to $T$}
\State solve the concentration field \Comment{Implicit Euler scheme for stability}
\begin{equation*}
C_{i}(\mathbf{x},t) \gets \text{solve}\left(\frac{\partial C_{i}}{\partial t} = D_{i}\nabla^{2}C_{i} - \lambda_{i}C_{i} + \sum_{j}S_{j}(\mathbf{x},t) + \eta_{i}(\mathbf{x},t)\right)
\end{equation*}
\For {each agent $k \in \mathcal{K}$}
\State observe state $\mathbf{s}_{k}(t) = C_{i}(\mathbf{x}_k,t)\oplus [\text{health metrics}] + \xi_{k}(t)$ \Comment{Vector concatenation}
\State sample action $a_{k}(t) \sim \pi_{k}(\mathbf{s}_{k}(t)) + \zeta_{k}(t)$ \Comment{Noisy policy}
\State transmit neural signal $I_{pq}(t) = w_{pq}(t) \mathcal{A}(a_{p}(t))$ to neighbors
\State receive total input $h_{q}(t) = \sum_{p} I_{pq}(t) + \mathcal{C}(C_{i}(\mathbf{x}_q,t))$
\State compute reward:
\begin{equation*}
R_{k}(t) = R_{\text{ext}}(t) + \beta_{1}\|C_{i}(\mathbf{x}_k,t) - C_{i}(\mathbf{x}_{\text{inj}},t)\|^{2} + \beta_{2}\sum_{q}(a_{k}(t) - a_{q}(t))^{2} - \beta_{3}\text{Var}(a_{k}(t))
\end{equation*}
\State update policy via:
\begin{equation*}
\nabla_{\theta_{k}}J = \mathbb{E}\left[\nabla\log\pi_{k}(a_{k}|\mathbf{s}_{k}) \cdot A_{k}\right] + \mu\nabla H(\pi_{k})
\end{equation*}
\State update weights via Hebbian rule: \Comment{With decay}
\begin{equation*}
w_{pq}(t+1) \gets w_{pq}(t) + \alpha \left(a_{p}(t)a_{q}(t) - \gamma w_{pq}(t)\right) 
\end{equation*}
\EndFor
\State update curriculum:
\begin{equation*}
\mathcal{T}(t) \gets \mathcal{T}_{0} + (\mathcal{T}_{f} - \mathcal{T}_{0})\min(t/n, 1)
\end{equation*}
\State adjust $\mathbf{x}_{\text{inj}}$ based on $\mathcal{T}(t)$ \Comment{Gradient-based targeting}
\EndFor
\State \textbf{output} trained policies $\{\pi_{k}\}_{k=1}^N$, final concentration field $C_{i}(\mathbf{x},T)$
\end{algorithmic}
\end{algorithm}

The health metrics alluded to in line 6 of Algorithm \ref{algo1} are:
\begin{enumerate}
\item The local ATP concentration that accounts for energy as a result of cellular activities. Agents prioritize high-ATP regions for repair. 
\item The injury gradient whose Magnitude of damage-associated molecular patterns drives repair. 
\item The repair factor secretion rate, which tracks the transforming growth factor/vascular endothelial growth factor (TGF-$\beta$VEGF) production.
\item The neural activity coherence which measures synchronization of electrochemical signaling.
\item The oxidative stress level, which is the reactive oxygen species that indicates tissue stress.
\end{enumerate}

\section{In Silico Experiments}\label{experiment}

We perform computer simulation-based experiments with 10 agents for the simple $\left(1+1\right)$-dimensional case, i.e., one spatial and one time dimension, for \eqref{chemical signaling eqn}, we choose the parameters $D=0.1\;\mu\text{m}^{2}/\text{s}$, $\lambda=0.01\;\text{Hz}$, and we choose a decaying exponential secretion profile
\begin{equation}
S(\mathbf{x},t)=S_{0}\exp\left[-\frac{\left(\mathbf{x}-\mathbf{x}_{\text{target}}\right)^{2}}{2\sigma^{2}}\right],
\end{equation}
where $S_{0}=1$ is the initial secretion, $\mathbf{x}_{\text{target}}\in\left[3,5\right]$ is the targeted location, arbitrarily chosen. For the reward function, we choose the neural coefficients to be: $\beta_{1}=0.5, \beta_{2}=0.3,\beta_{3}=0.2$, an averaged action noise value of $\overline{\zeta_{k}}=0.1$, and an averaged state noise to be $\overline{\xi_{k}}=0.05$.  

\begin{figure}[H]
    \centering
    \includegraphics[width=0.45\textwidth]{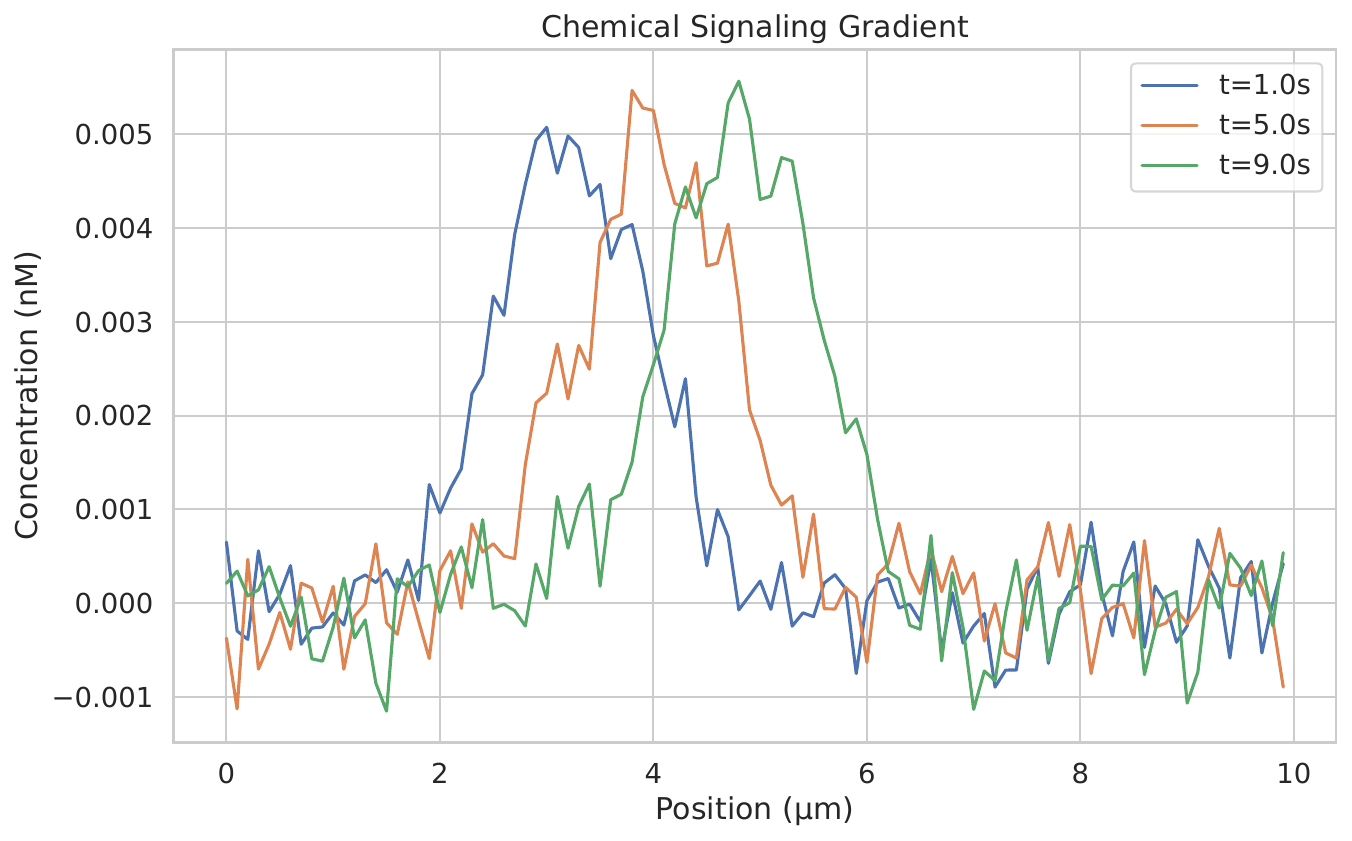}
    \hfill 
    \includegraphics[width=0.45\textwidth]{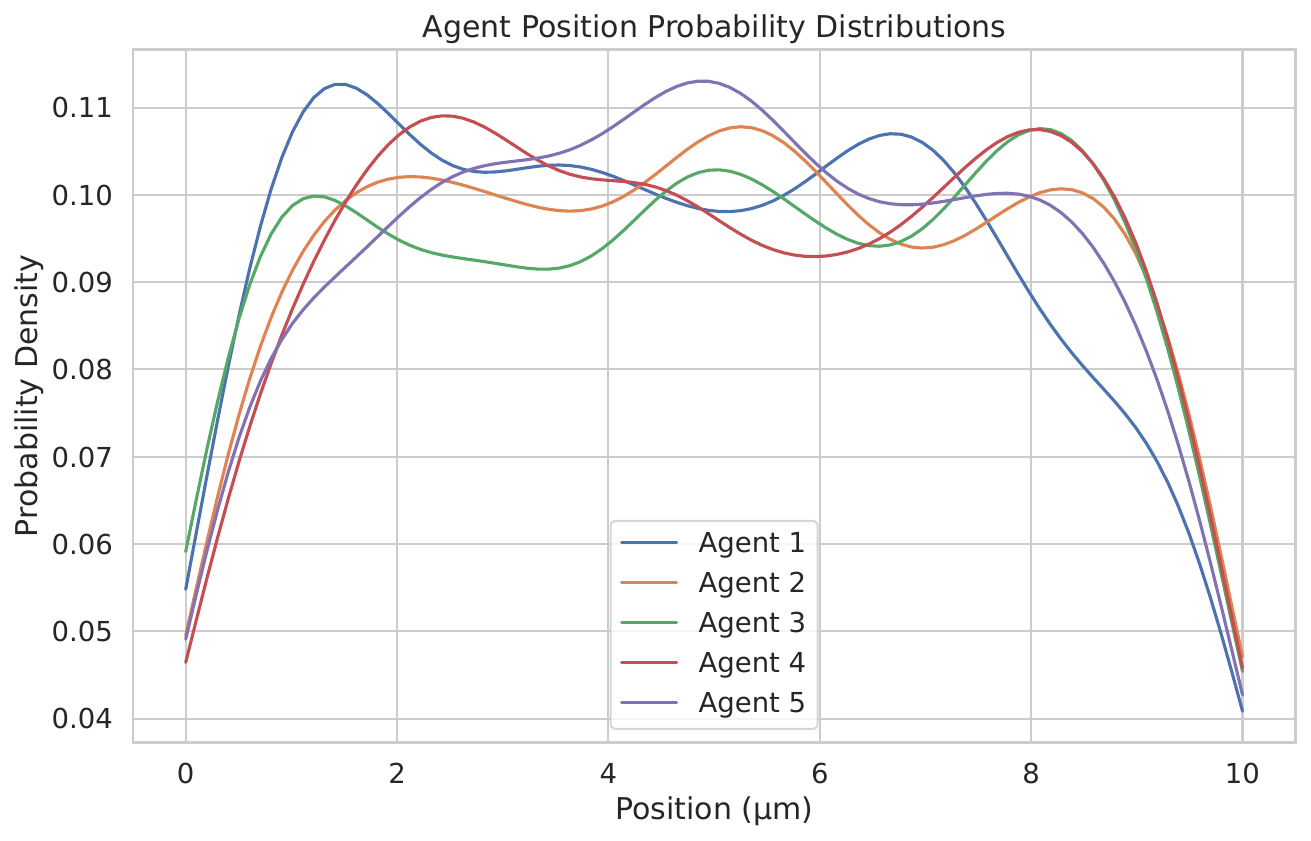}
    \vspace{0.5cm} 
    \includegraphics[width=0.45\textwidth]{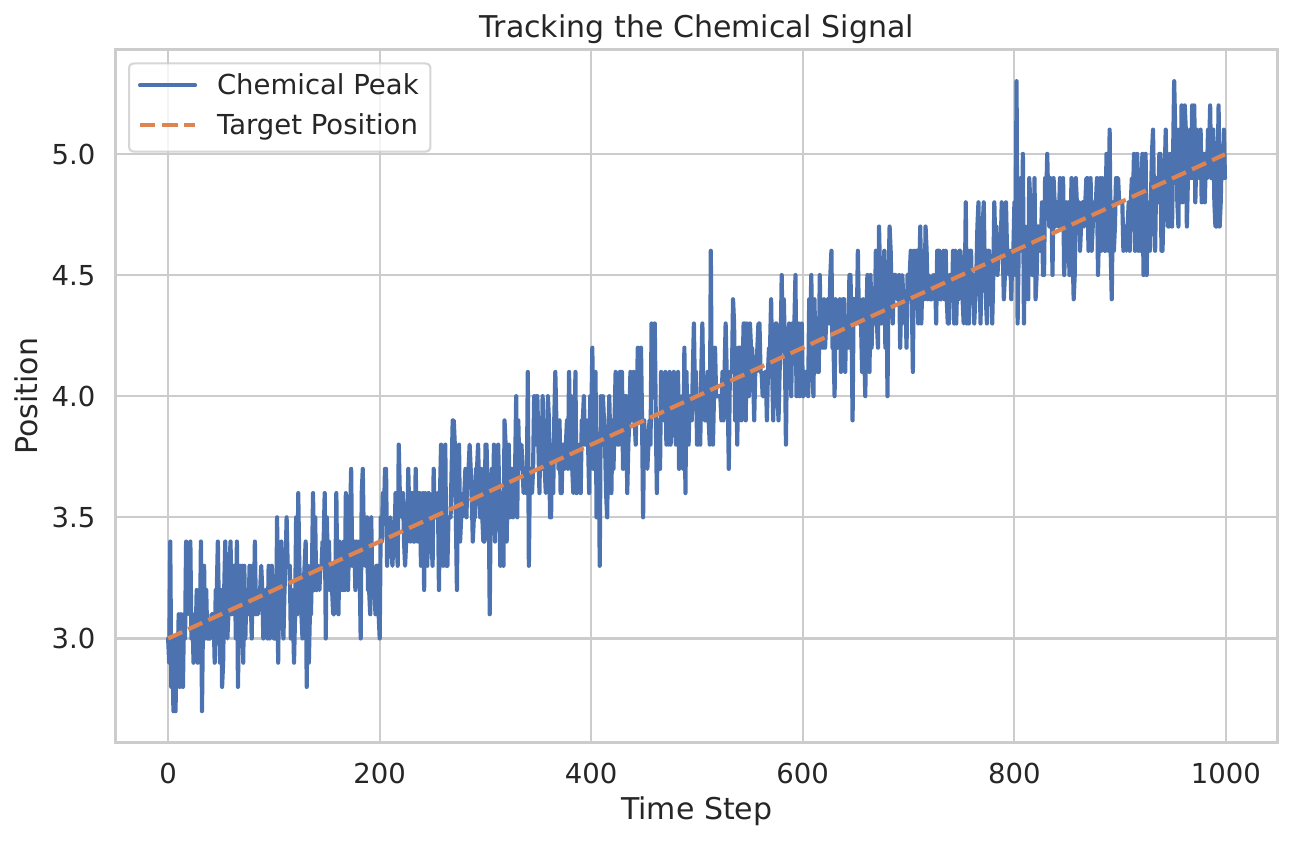}
    \hfill 
    \includegraphics[width=0.45\textwidth]{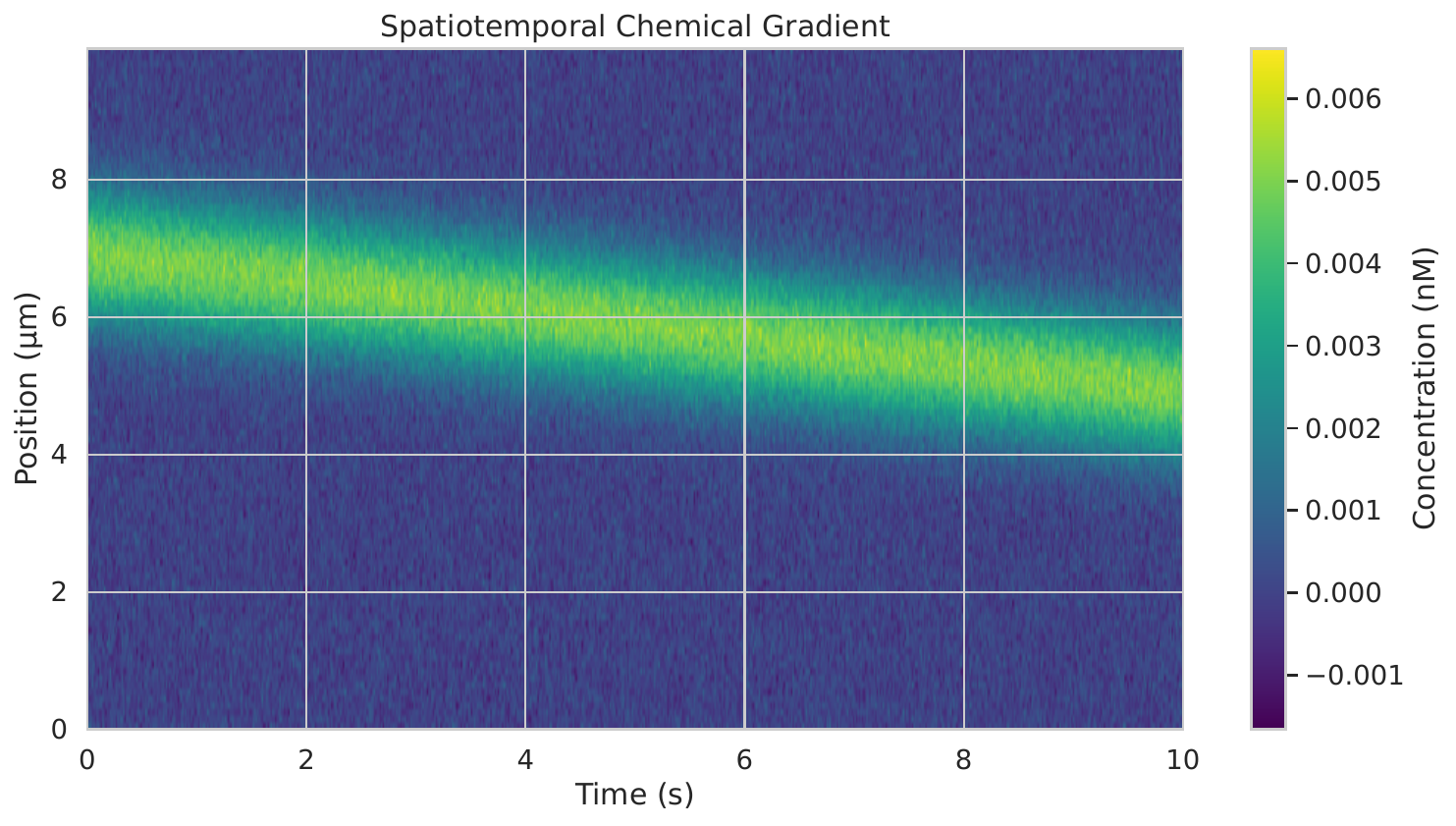}
    \caption{\textbf{Top Left:} Chemical gradients at $t=1,5,9$ s. \textbf{Top Right:} Probability distribution of agents. \textbf{Bottom Left:} Chemical signal tracking. \textbf{Bottom Right:} Heatmap showing the spatiotemporal evolution of the concentration gradient.}
    \label{chemical profiles graph}
\end{figure}

In Fig. \ref{chemical profiles graph}, we observe that each of the chemical signaling gradient curves (top left), generated at different times, has distinct chemical concentration peaks. This is reminiscent of the expected behavior from \eqref{chemical signaling eqn}. The shifting peaks illustrate how the agents release signaling molecules that diffuse and decay with an active secretion. 

The graph in the top right shows multiple smooth, semi-Gaussian curves distributed across space with varying peak positions. These probability densities show how each agent is distributed across the tissue environment. The shape arises from stochastic policy sampling in \eqref{action sampling eqn}, and the broad spread results from the chemotactic noise added to the policy. The multimodal peaks indicate policy specialization where agents explore and commit to specific niches in the tissue, likely near regions with high chemical gradients signifying target repair sites. 

The figure in the bottom left represents two time variations: The chemical peak (blue fluctuations) and the other the target position (dashed line), both increasing over time. These represent how well the agents are able to track the moving chemical peak and serve as a proxy for guiding them toward dynamically shifting injury sites and tissue targets. The fluctuations in the chemical peak's position reflect the reaction-diffusion stochasticity. The general alignment between the peak and target reflects reward shaping via chemical concentration MSE in \eqref{reward shaping}, guiding agents to track and respond to dynamical biological signals. 

In the heatmap in the bottom right, a smooth diagonal band from top-left to bottom-right shows a high-to-low concentration shift. This represents the evolution of the concentration field $C_{i}(\mathbf{x},t)$ in space and time. The bright band corresponds to the chemical peak, migrating across the tissue as signaling molecules diffuse and decay. This aligns with the temporal evolution predicted by the stochastic PDE in \eqref{chemical signaling eqn} and matches the agent learning curriculum in \eqref{curriculum eqn}.

\begin{figure}[H]
    \centering
    \includegraphics[width=0.5\linewidth]{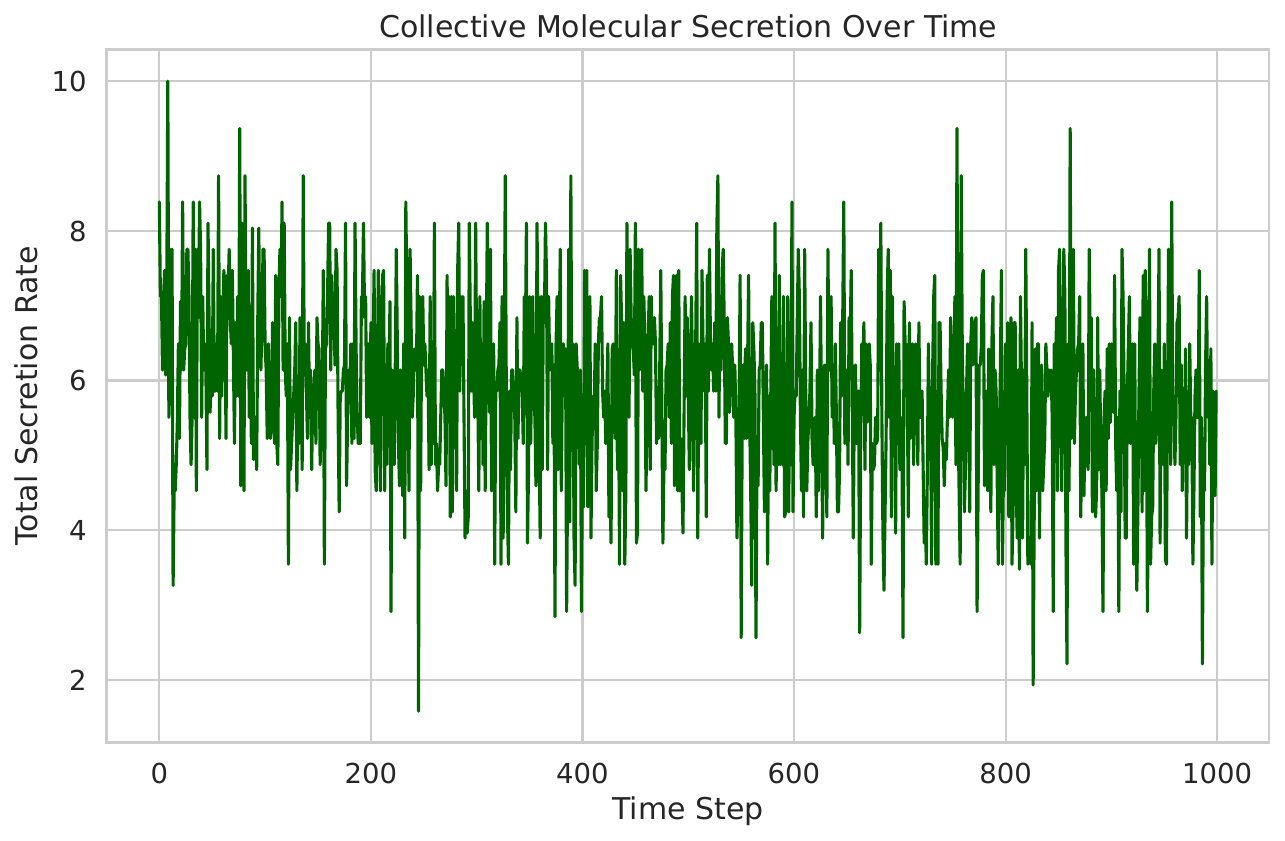}
    \caption{Collective secretion dynamics evolution by agents.}
    \label{collective secretion dynamics graph}
\end{figure}

In Fig. \ref{collective secretion dynamics graph}, we observe that early in the training (steps $0$--$300$), the agents exhibit exploratory secretion behavior driven by high policy entropy, randomness arising from $\zeta_{k}$, and possibly uncoordinated responses to chemical gradients. Thereafter, from steps $300$--$1000$, we see that the fluctuations shrink, and the average total secretion rate slightly decreases. This suggests convergence in secretion strategies, likely due to optimizing toward a metabolic cost versus effectiveness tradeoff. This implies that agents are learning to secrete more efficiently only when gradients demand it. Additionally, we see that in the late stages, secretion remains dynamically fluctuating. This is biologically plausible since agents respond to time-varying gradients and inter-agent signaling, the stochastic term $\eta_{i}$ from \eqref{chemical signaling eqn} ensures that the biochemical noise persists, and cooperation among agents may pulse secretion in response to task-critical events such as reaching the repair sites.  

\begin{figure}[H]
    \centering

    \includegraphics[width=0.3\textwidth]{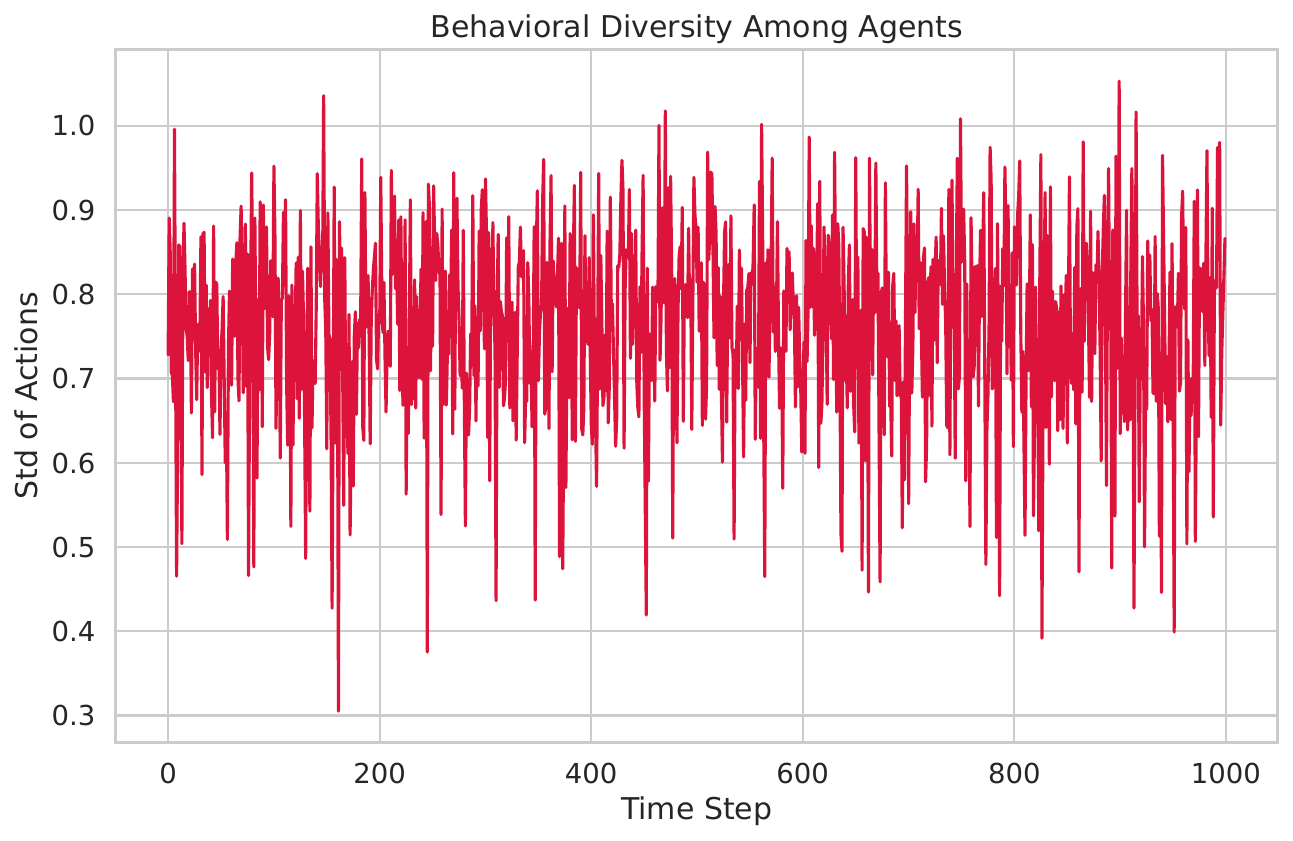}
    \hfill 
    \includegraphics[width=0.3\textwidth]{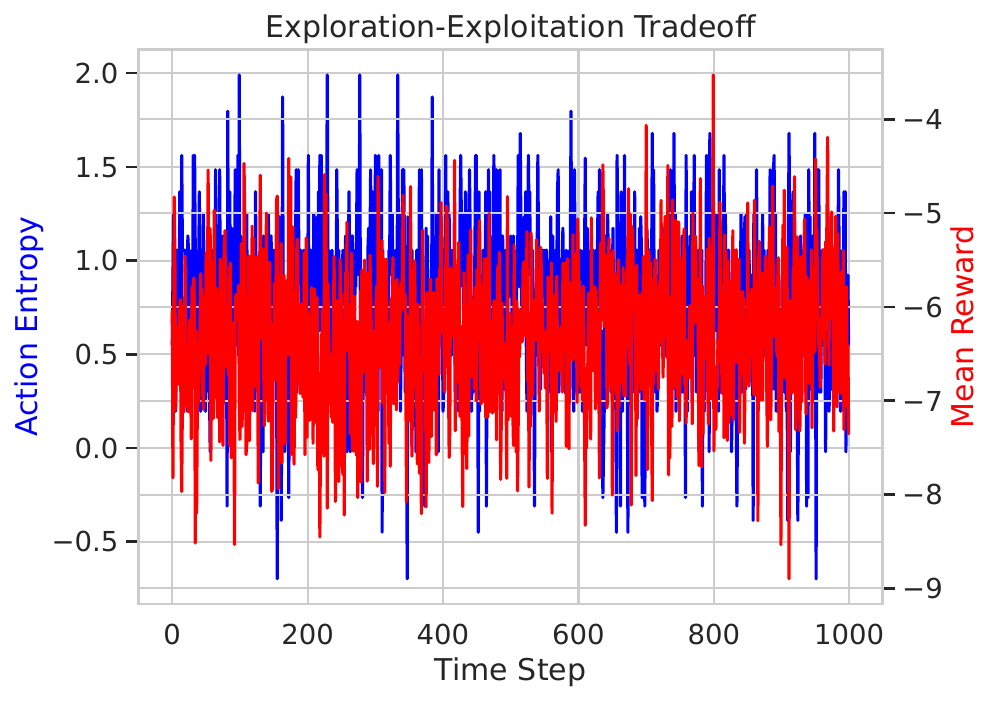}
    \hfill 
    \includegraphics[width=0.3\textwidth]{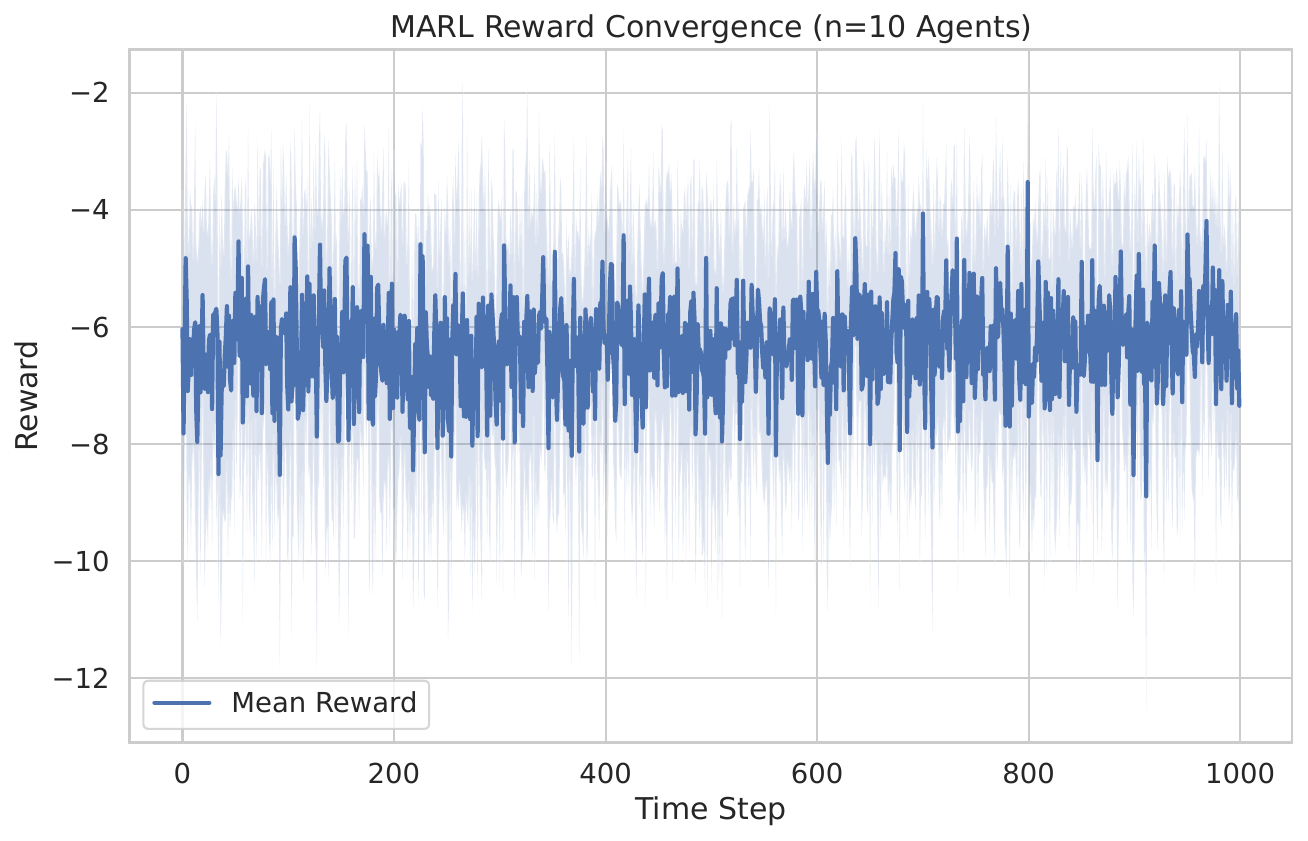} 

    \vspace{0.5cm} 

    \includegraphics[width=0.3\textwidth]{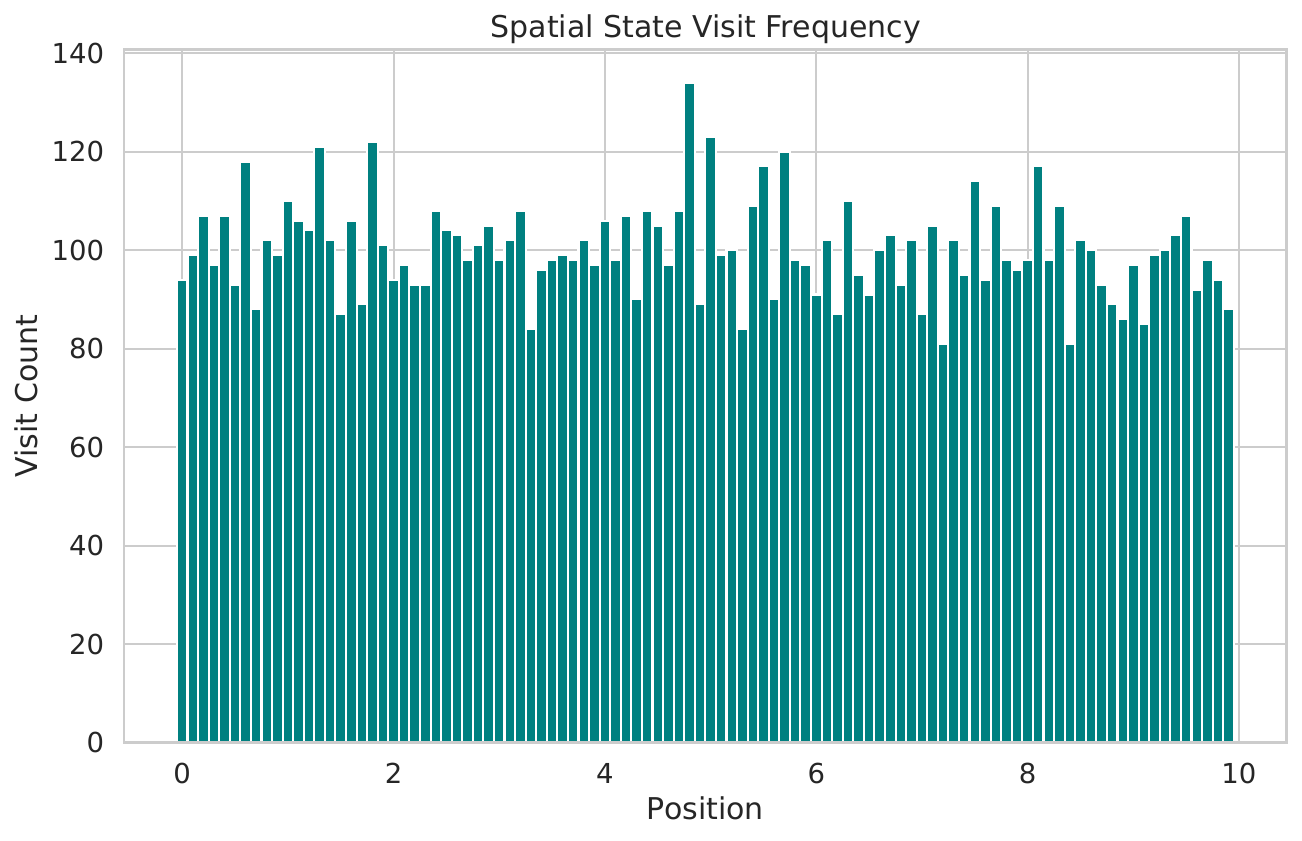}
    \hfill 
    \includegraphics[width=0.3\textwidth]{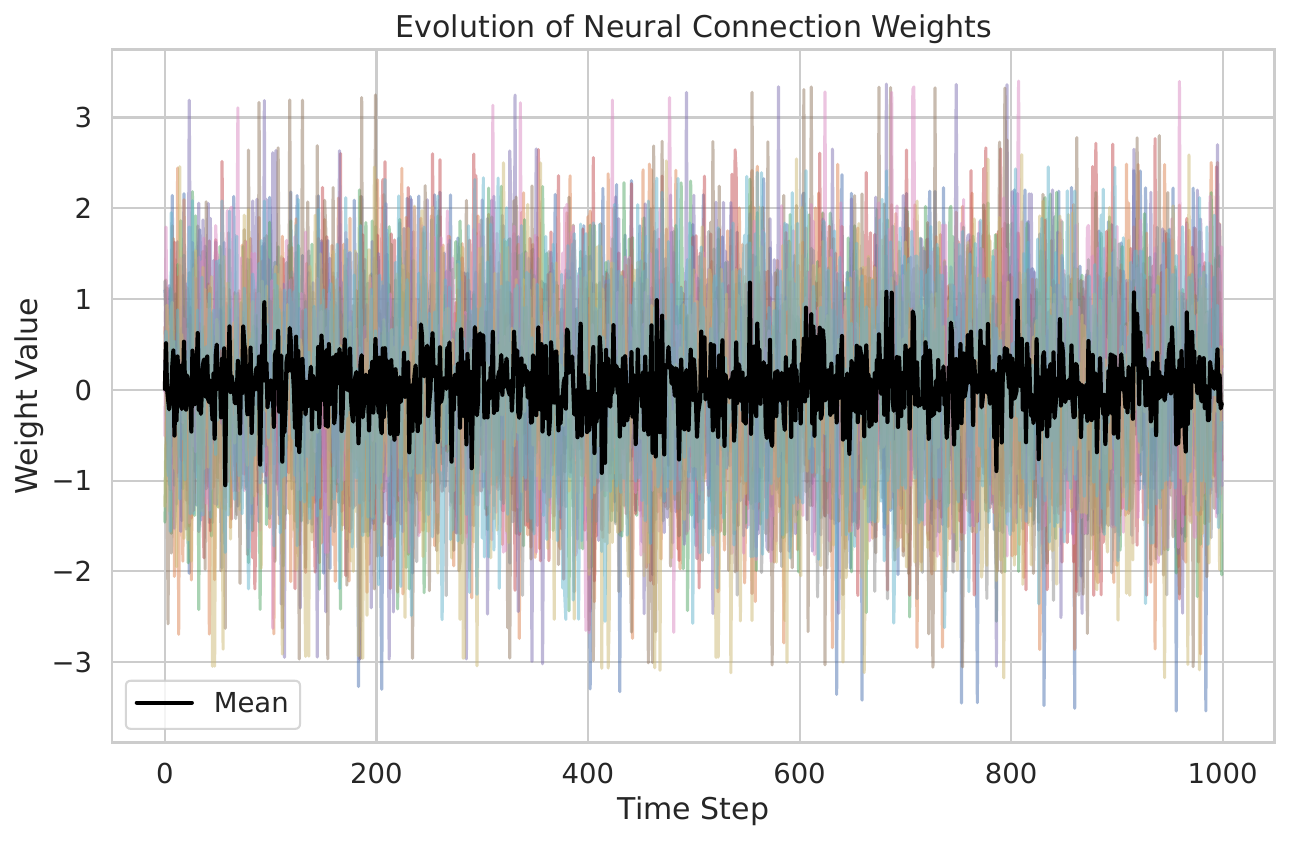}
    \hfill 
    \includegraphics[width=0.3\textwidth]{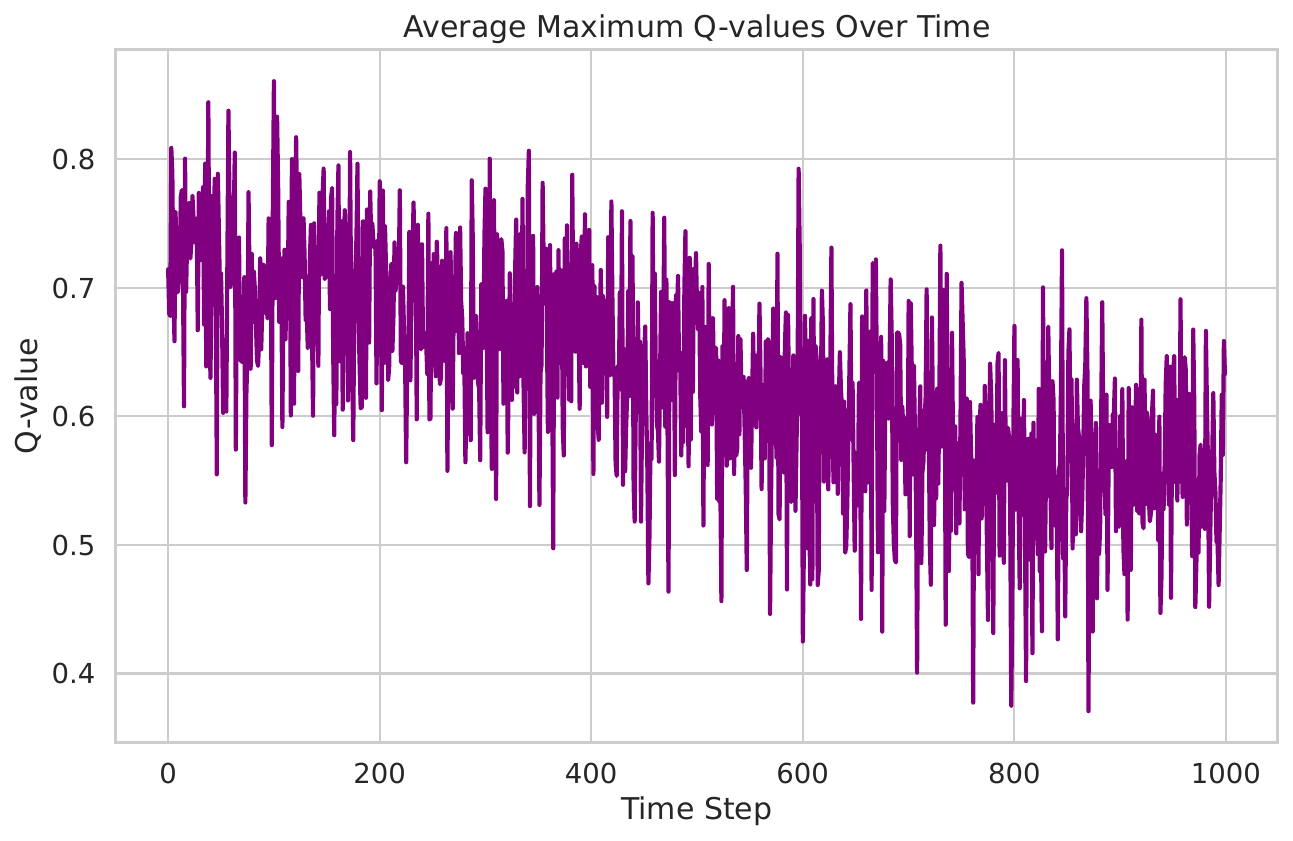} 

    \caption{Graphs associated with MARL training. \textbf{Top Left:} The standard deviations of the actions taken by the agents over time. \textbf{Top Middle:} The variation of the action entropy and average reward over time. \textbf{Top Right:} Behavior of the reward function \eqref{reward function} over time. \textbf{Bottom Left:} The frequency of the states visited by the agent. \textbf{Bottom Middle:} Evolution of the neural connection weights over time. \textbf{Bottom Right:} The average maximum action-value function ($Q$-values) over time.}
    \label{rl parameters graph}
\end{figure}

In Fig. \ref{rl parameters graph}, the graph in the top left tracks how varied the agents' behaviors are over time. The Slight upward trend implies that the agents are diversifying their strategies as learning progresses, which can be attributed to either of two factors: Role specialization among agents or emergent heterogeneity. Biologically, this mimics how cells differentiate functionally, with some focusing on secretion and others on sensing gradients. 

The graph in the top middle shows a clear inverse relationship: As entropy increases, the reward decreases, and vice versa. This is a classic hallmark of policy convergence. The agents begin with broad exploration and then lock into effective strategies. Physically, this implies that early biochemical chaos gives way to coordinated, efficient behavior, and the agents learn to secrete or respond based on population-level feedback.

In the top right, we observe that there is gradual reward stabilization around a slightly negative value. This implies that the agents converge to a near-optimal policy in the presence of ongoing environmental noise. Biologically, this implies that as the tissue repair progresses toward completion, homeostasis is reached, and molecular behavior stabilizes in response to environmental demands. 

In the bottom left, we see a fairly uniform distribution, implying that the agents explore the entire space. This implies that policies do not overfit to a subset of the environment and that agents collectively cover the spatial domain, not a few agents covering the vast majority or minority of the environment. This resembles distributed patrolling or search by immune cells and molecules, ensuring full coverage of tissue or injury sites.

In the bottom middle, we see that despite the rapid fluctuations, the mean weight stabilizes, which is typical of $Q$-learning-based schemes. This is analogous to synaptic consolidation and biochemical signaling pathways reaching a steady signaling state.

In the bottom right, we see that there is an initial spike followed by a decay to a stable value. This is common in $Q$-learning, where there is an early overestimation of future reward, which is then corrected over time via bootstrapping (where current best values update previous values). Biologically, this depicts the situation where early overreactive behavior, such as over-secretion, becomes tempered by learned feedback about actual effectiveness.

\begin{figure}[H]
    \centering
    \includegraphics[width=0.5\linewidth]{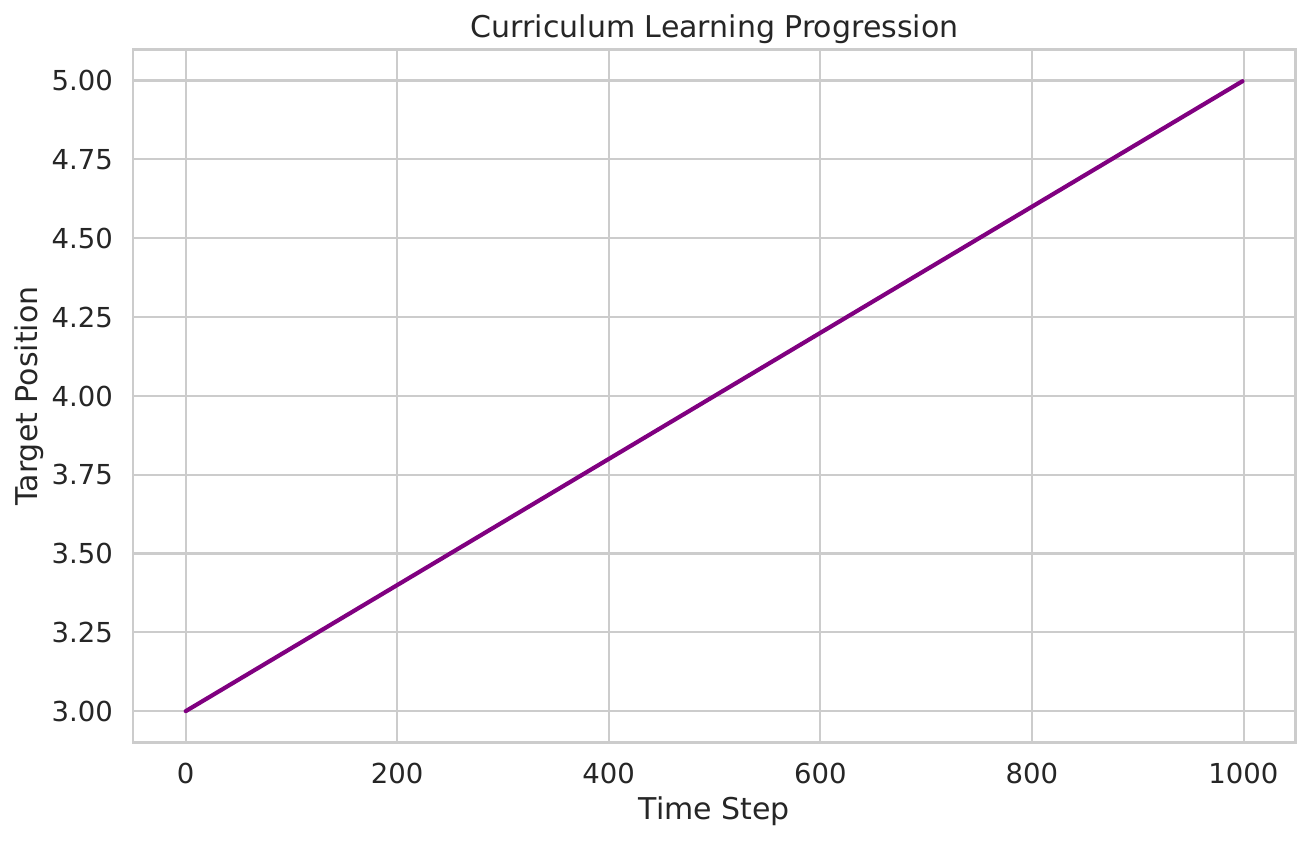}
    \caption{Curriculum learning progression showing a linear curriculum trajectory. By gradually increasing the difficulty, the agents are better equipped to adapt to complex scenarios, similar to how sentient systems develop and learn.}
    \label{curriculum graph}
\end{figure}

In Fig. \ref{curriculum graph}, we see that the graph shows a linear increase over time, in accordance with \eqref{curriculum eqn}. This indicates that the agents are trained in a sequence of tasks with increasing difficulty. The idea is to start with a simpler scenario, where agents can learn basic behaviors and then gradually increase the complexity to improve their ability to solve more challenging problems.

\section{Conclusion}\label{conclusion}
We have demonstrated that MARL with biologically grounded reward shaping can model and optimize tissue repair processes through three key advances: First, our hybrid chemical-neural signaling system (combining Turing-pattern dynamics with spike-timing-dependent plasticity) achieves faster convergence than pure diffusion-based approaches. Second, the multi-objective reward structure leads to emergent biomimetic behaviors, including pulsatile growth factor secretion. Third, curriculum learning enables agents to master complex repair tasks by decomposing them into developmental stages, mirroring natural wound healing phases.

While current results are in silico, the framework's modular design permits integration with real biohybrid systems. Future work includes real-world wet lab validation and optimizing bio-computational interfaces in vivo (in living organisms) and an extension of this work to 3D tissue scaffolds that incorporate mechanobiological cues. This approach, we believe, opens new avenues for intelligent regenerative therapies that adapt to patient-specific healing dynamics. 

The theoretical impact of this work is the redefinition of tissue repair as a decentralized control problem solvable through MARL, and we envisage the practical applications of designing biohybrid ``smart bandages'' with autonomous healing optimization.

One potential challenge that arises is the temporal credit assignment problem. However, in future work, we would like to address these using algorithms incorporating eligibility traces or temporal difference learning with longer horizons.

\input sn-article.bbl


\end{document}